\documentclass[conference,a4paper]{IEEEtran}
\pdfoutput=1
\usepackage{epsfig}
\usepackage{graphicx}
\usepackage{amsmath}
\usepackage{amssymb}
\usepackage{multirow}

\begin{document}
\title{Overlay Text Extraction From TV News Broadcast}
\author{\IEEEauthorblockN{Raghvendra Kannao and Prithwijit Guha} 
\IEEEauthorblockA{Department of Electronics and Electrical Engineering, IIT Guwahati\\ Assam, India 781039\\
	Email: [raghvendra, pguha]@iitg.ernet.in}}

\maketitle
\begin{abstract}
The text data present in overlaid bands convey brief descriptions of news events in broadcast videos. The process of text extraction becomes challenging as overlay text is presented in widely varying formats and often with animation effects. We note that existing edge density based methods are well suited for our application on account of their simplicity and speed of operation. However, these methods are sensitive to thresholds and have high false positive rates. In this paper, we present a contrast enhancement based preprocessing stage for overlay text detection and a parameter free edge density based scheme for efficient text band detection. The second contribution of this paper is a novel approach for multiple text region tracking with a formal identification of all possible detection failure cases. The tracking stage enables us to establish the temporal presence of text bands and their linking over time. The third contribution is the adoption of Tesseract OCR for the specific task of overlay text recognition using web news articles. The proposed approach is tested and found superior on news videos acquired from three Indian English television news channels along with benchmark datasets.
\end{abstract}


%

\section{Introduction}
\label{sec:intro}

Overlay text bands in TV broadcast news videos provide us with rich semantic information about news stories which are otherwise hard to estimate by processing audio-visual data. In Indian TV news broadcast, overlay text has widely varying formats and occupies $20-40\%$ of the total screen area (during regular news presentations and increases while presenting headlines etc.). Hence, overlay text is an important feature in a number of sub-tasks of news video analysis viz. news summarization, story segmentation, indexing and linking, commercial detection etc.

A video text extraction pipeline generally involves text detection and localization in each frame, text tracking over the frames and recognizing the text using an OCR engine. In TV news broadcast, text is overlaid in the form of different text bands. Text bands contain single or multi-line sentences or semantically linked set of words (e.g. name of person followed by designation in next line). Different text bands have different semantic meanings and are characterized by on screen position and style of the text band. For example, text relevant to a story is often overlaid in upper part of the screen in large font size and with high contrast in colors. Thus, instead of identifying discrete words, we propose to detect, track and recognize text from \textbf{overlay text bands}. We define a text band as a region bounded by a rectangle enclosing one or more adjoining text regions (words) subject to following conditions. First, all the text regions should have almost same stroke width. Second, no sharp changes in background as well as foreground (text) color of text regions. Third, all the text regions should have common base line. Fourth, text regions should not have any separator between them.

Overlay text detection schemes can be categorized as either patch based or geometrical property based \cite{SWT:CVPR2010}. Patch based techniques extract features from image patches and identify text regions using pre-trained classifiers \cite{minetto:ICCV2011}. These patches are grouped further to detect the text regions. These methods have shown excellent performance on various challenging real life problems but at the cost of rigorous pre-training. On the other hand, geometrical property based methods make assumptions on representative features of text regions like high edge/corner density \cite{Rosenfeld/Azriel:2003,Tang/Luo:ICME2002}, edge continuity \cite{Koo/Kim:2013}, stroke consistency \cite{SWT:CVPR2010,Yao/Bai:CVPR2012}, and color consistency \cite{JainYu:PR1998}. These methods are well known for off the shelf deployment (as no pre-training is required) and fast speed of operation.

The simplest of the geometrical property based approaches rely on edge or corner densities. Edge/corner density based methods assume high density of strong edges in text regions and have been used extensively due to their simplicity and speed \cite{Tang/Luo:ICME2002}. However, these approaches require selection of thresholds and suffers due to high edge density in non-text regions resulting in false positives. Based on different properties of text regions, various curative measures were proposed in literature in order to suppress strong edges from non-text regions and hence, the false positives \cite{lyu:csvt2005}. For example, Kim et.al \cite{wkim/:IP2009} have observed that text regions have peculiar color transition pattern due to high contrast and hence, suggested the use of color transition maps instead of edge images to calculate the density.

We believe that for comparatively simple task of overlaid text detection using patch based approaches \cite{minetto:ICCV2011} and some of the complex methods like stroke width transform \cite{SWT:CVPR2010} will overkill the resources. In this work we build up on basic edge density based text detection and propose the following to improve the video text extraction pipeline typically for TV news broadcast videos. First, we propose to reduce false positives in edge density based methods by selectively boosting overlaid text edges using contrast enhancement based preprocessing scheme. Moreover, we use derivatives of edge projection profiles for threshold free detection of text bands. Second, we propose a spatial relations based reasoning framework for tracking multiple (detected) text bands, which is capable of identifying and handling various problems arising out of detection/tracking failures. Third, we propose to improve the performance of Tesseract OCR engine by using synthetically generated training data and incorporating a dictionary of words derived from web news articles. The rest of this paper is organized as follows. The methodology for frame-wise text band detection is presented in Section~\ref{sec:txtDet}. The proposed approach for multiple text region tracking in videos is presented in Section~\ref{sec:mtt}. Modifications in Tesseract OCR engine to improve the recognition rate are described in Section~\ref{sec:ocr}. The experimental results are discussed in Section~\ref{sec:res}. Finally we conclude in Section~\ref{sec:conc} and sketch the future scope of work.

\section{Text Band Detection}
\label{sec:txtDet}

Overlay text bands have usually clutter free background, high contrast between foreground (text) and background and doesn't suffer from perspective distortions. Hence, comparatively faster and easy to deploy geometrical approaches are a natural choice for overlaid text detection. We have adopted and improved on a well established edge density based approach \cite{Rosenfeld/Azriel:2003} for detecting text bands instead of discrete words. The basic edge based method detects and localizes text regions by using horizontal and vertical projection profiles of gradient magnitude (edge) image.  Performance of the basic method deteriorates due to high edge densities in non-text regions, mis-alignment of different text bands and high sensitivity to thresholds. We propose to reduce the false positives by a preprocessing technique to selectively boost text edges while suppressing non-text edges based on contrast of gradient magnitude image. We use first and second derivatives of projection profiles to reduce the dependence on projection profile thresholds. Our proposed preprocessing scheme is described next.

\begin{figure}[htbp]
\centerline{
\includegraphics[width=0.4\columnwidth]{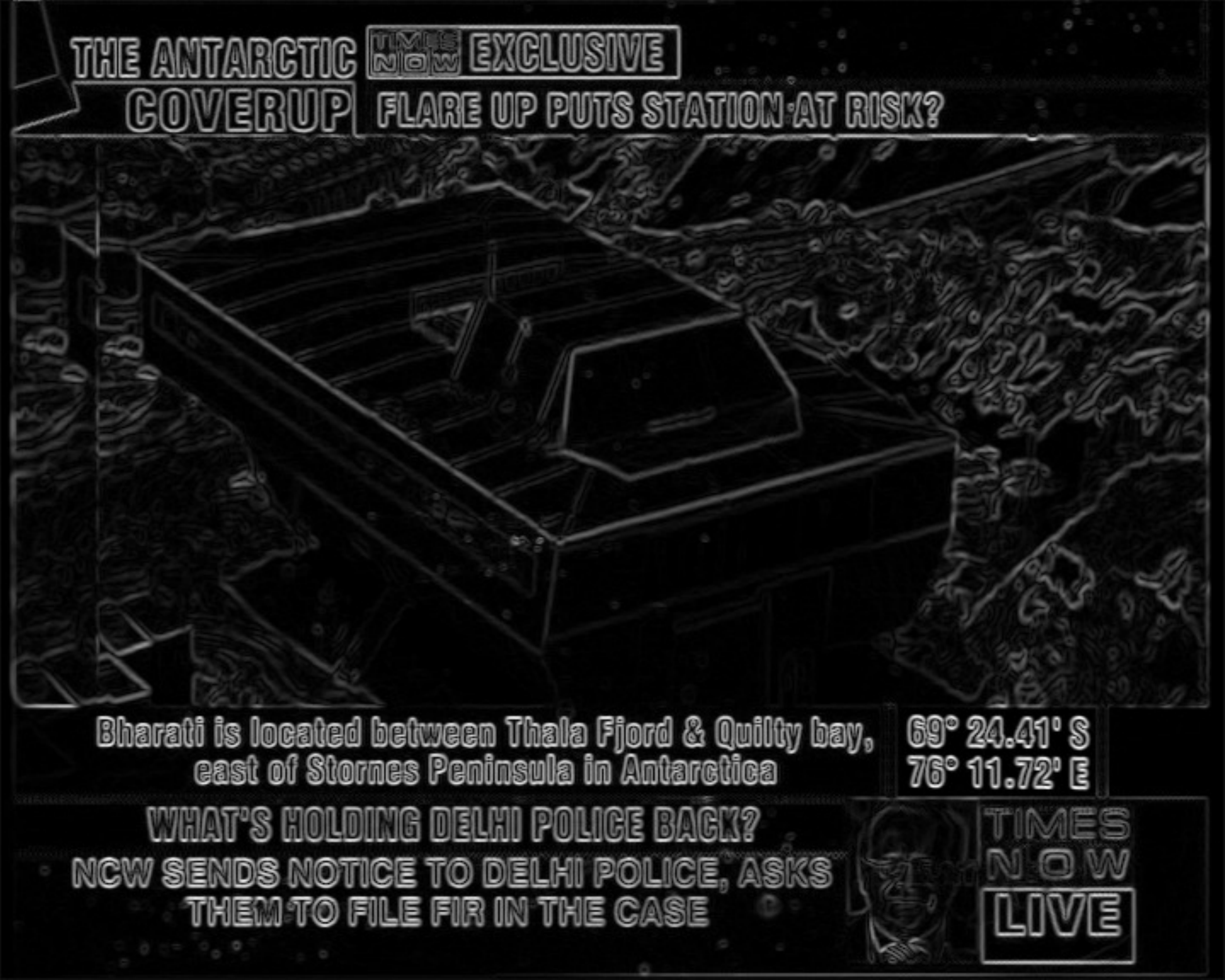}
\includegraphics[width=0.4\columnwidth]{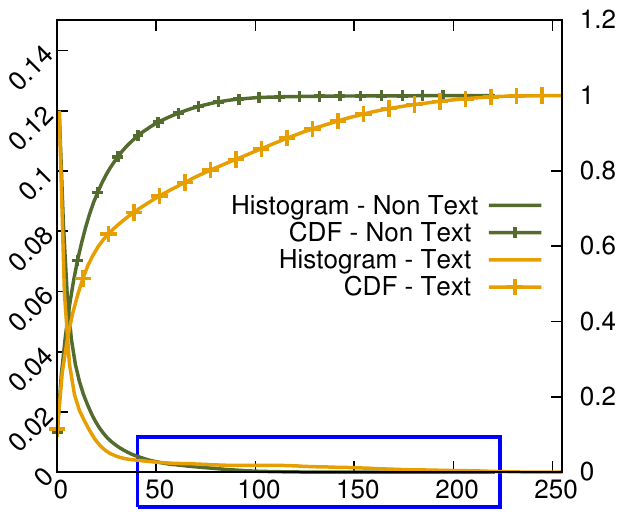}
}
\centerline{(a)\hspace{0.4\columnwidth}(d)}
\centerline{
\includegraphics[width=0.4\columnwidth]{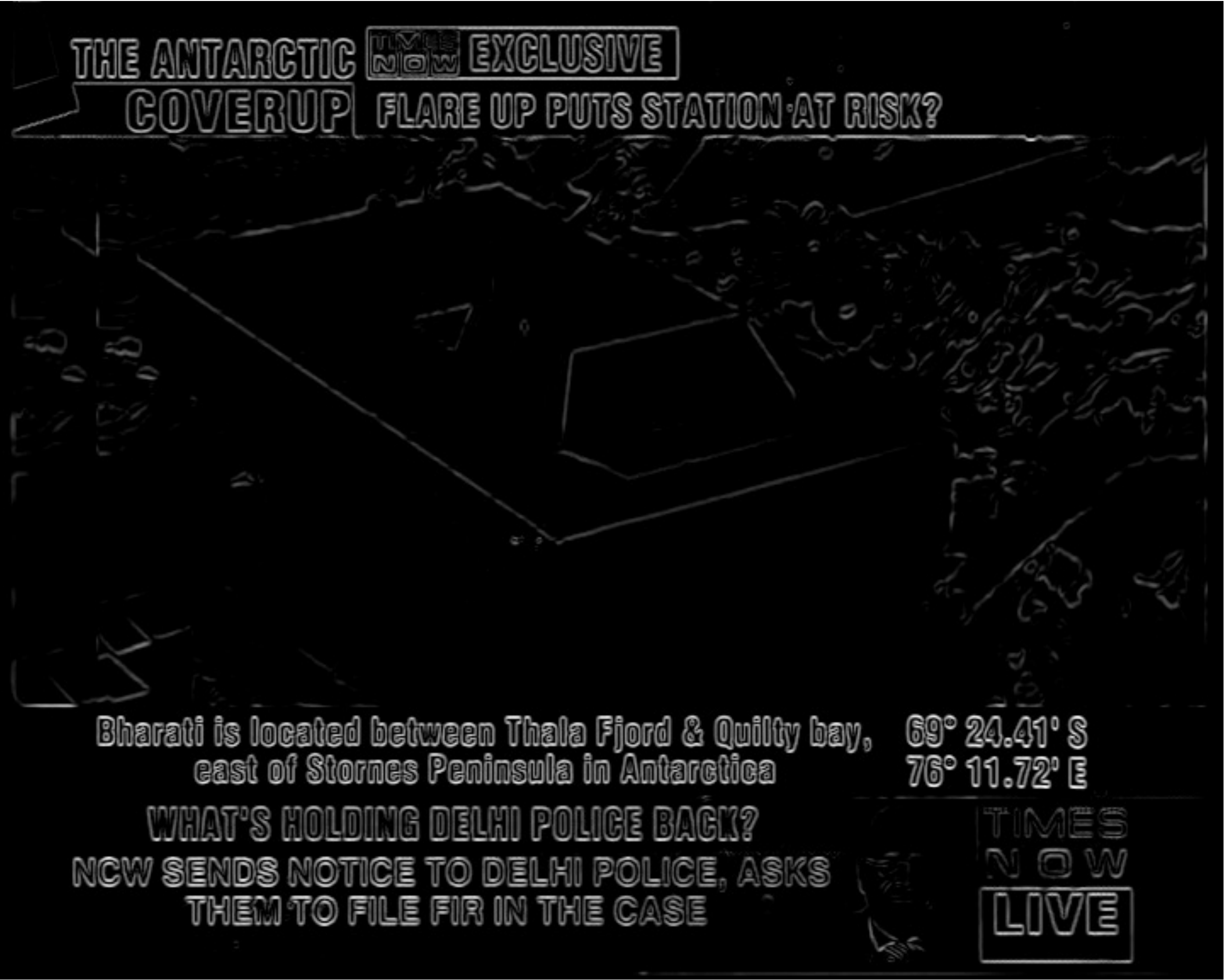}
\includegraphics[width=0.4\columnwidth]{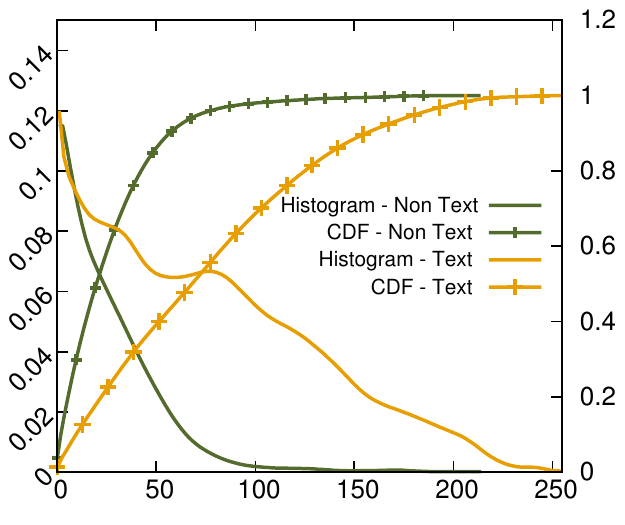}
}
\centerline{(b)\hspace{0.4\columnwidth}(e)}

\centerline{
\includegraphics[width=0.4\columnwidth]{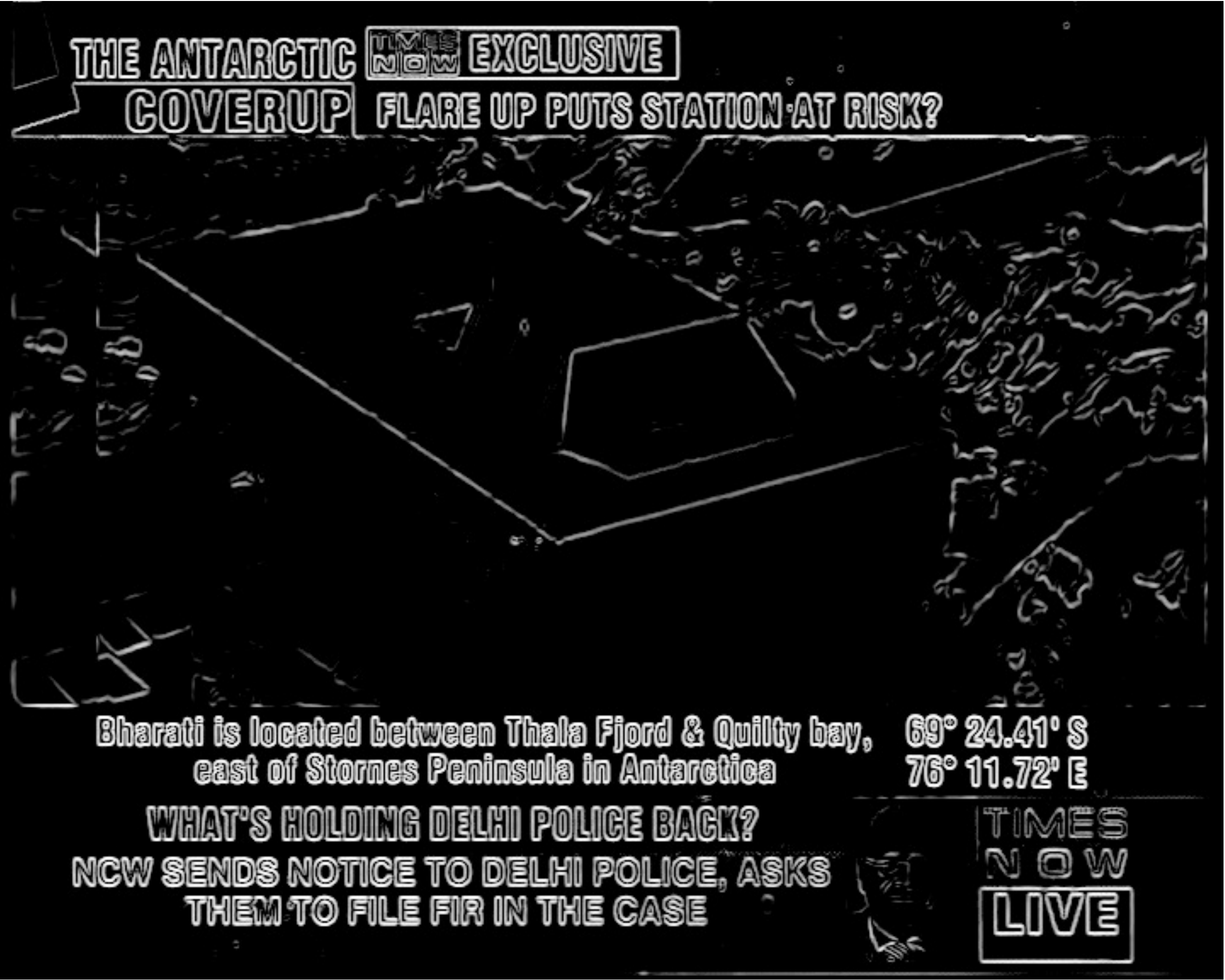}
\includegraphics[width=0.4\columnwidth]{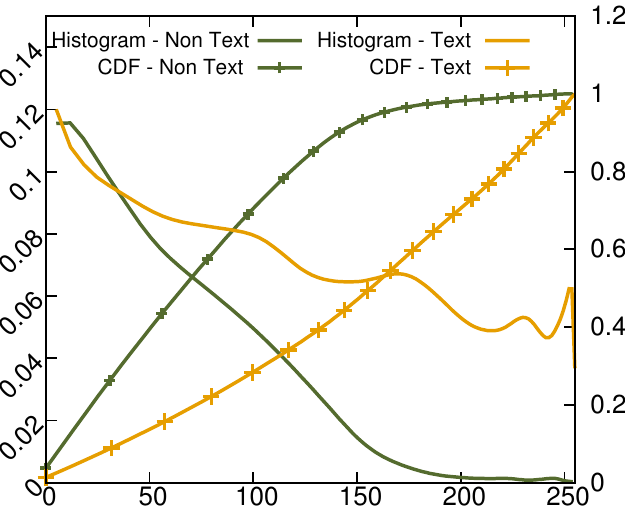}
}
\centerline{(c)\hspace{0.4\columnwidth}(f)}
\caption{\footnotesize{Figure (a) - (c) shows gradient magnitude image, gradient magnitude image after linear contrast stretching and gradient Magnitude image after histogram equalization, respectively. Figures (d)-(f) shows the gradient value histograms and Cumulative Distribution Functions of images (a)-(c) . Overlay text regions have comparatively strong gradients due to high contrast as evident from (a) and (d). We selectively boost the stronger gradients so as to have sufficient separation between text and non-text gradient values. This boosting is achieved first by stretching of histogram (Figure (d)) followed by histogram equalization (Figure (f)).  Note the steep change in CDF of gradients from text and non-text regions in histogram (f). Hence, after pre-processing text regions ($P(x_{text} < 150) = 0.45$) can now be easily discriminated from non-text regions ($P(x_{nontext} < 150 )) = 0.95$).}}
\vspace{-2em}
\label{fig:preProc}
\end{figure}

\subsection{Preprocessing Proposal}
\label{subsec:preProc}

In edge density based text detection, the first step is to calculate gradient magnitude/edge image followed by binarization of gradient magnitude image to get an edge map. Threshold for binarizing the gradient magnitude is selected such that, edge map should have edges only from text regions. This binarization threshold is a critical parameter for basic text detection approaches \cite{wkim/:IP2009,Tang/Luo:ICME2002,lyu:csvt2005}. However, gradient magnitudes from text and non-text regions have significant overlap (Figure~\ref{fig:preProc}(d)) and hence, choosing a threshold to discriminate text and non-text gradients is not trivial. In cases of such overlaps, usually a lower threshold is chosen to eliminate definitely non-text regions. Further, local text specific features are used to eliminate strong edges from non-text regions \cite{lyu:csvt2005}.

Due to high contrast of text bands, gradient magnitudes originating from text regions lie on comparatively higher side of the edge magnitude histogram (Figure~\ref{fig:preProc}(d)). We propose to increase the dynamic range of gradient magnitude histogram to increase discrimination between text and non-text gradient magnitudes. We achieve this in the following two steps. First, by linearly stretching (linear contrast enhancement) the gradient magnitude histogram and second, by equalizing the stretched gradient histogram. We use isotropic Scharr operator \cite{lyu:csvt2005} to compute the gradient magnitude image over the Sobel operator proposed in basic method \cite{Rosenfeld/Azriel:2003}. We start with normalizing the gradient magnitude image $I_{m}$ by the maximum gradient magnitude value ($g_{max}$) to obtain the image $I_{nm} = I_{m}/g_{max}$. Next, we eliminate edges from definitely non-text regions having small gradient magnitudes by linearly stretching gradient magnitude histogram (Figure~\ref{fig:preProc}(e)). The gradient magnitude image after linear stretching is, $ I_{mc}(x,y) =  \beta / \lambda$ where, $\beta = \alpha( I_{nm}(x,y) - 0.5 ) + 0.5$ and $\lambda = max( I_{mc} )$ is the normalizing constant. The factor $\alpha$ ($\alpha > 1$) decides the extent of suppression. The value of $\alpha$ should be selected so as to nullify gradient magnitudes from ``definitely non-text regions''. The lowest non-suppressed gradient magnitude value is $g_{ns} = \frac{ (\alpha - 1) g_{max}}{ 2 \alpha}$ where, $g_{max}$ is the maximum gradient value. Thus, the value of $\alpha$ is not independent of data and we select its value using Otsu's thresholding scheme. Otsu's threshold gives upper bound on values of gradient magnitudes originating form definitely non-text regions. We select value of $\alpha$ such that, $g_{otsu} = g_{ns}$ or $\alpha = (g_{max})/( g_{max} - 2 g_{otsu})$. The final expression for  stretched histogram image (Figure~\ref{fig:preProc}(b)) is given by, $I_{mc}(x,y) = \frac{ u( I_{mc}(x,y) - g_{ns})}{\lambda}\left[ \alpha( I_{nm}(x,y) - 0.5 ) + 0.5 \right]$ where, $u(\cdot)$ is the unit step function.

Histogram equalization is done further on $I_{mc}$ to obtain the edge map $\Omega_{ce}$ which we use further for text band detection. The histogram equalized image (Figure~\ref{fig:preProc}(f)) now have well distributed gradient magnitudes. Also, it is to be noted that gradient magnitudes originating from text regions are concentrated towards higher side of the pixel value histogram, while non-text magnitudes are concentrated towards lower side. Hence, the preprocessing stage successfully suppresses false positives originating due to strong non-text edges. Our next proposal for threshold free overlaid text band detection using derivatives of horizontal and projection profiles is described next (Sub-section~\ref{subsec:txtBandDet}).

\subsection{Text Band Detection}
\label{subsec:txtBandDet}

The Basic method of text detection proposed in \cite{Rosenfeld/Azriel:2003} assumes high edge density in text regions. The basic method uses horizontal and vertical projection profile of gradient magnitude or edge image to locate high edge density text regions. Text bands are aligned horizontally. Thus, horizontal projection profile is processed first to obtain bands having sufficient edge density. Horizontal projection profile $P_{hp}$ of edge image $\Omega$ of size $w \times h$ is given by, $P_{hp}(y) = \sum_{x = 1}^{w} \Omega(x,y)$, where $y = 1,\ldots h$. Edge profile $P_{hp}$ is thresholded by a threshold $\eta_{hp}$ to discard horizontal lines having insufficient edge density. A region between $y_{i}$ and $y_{j}$ is marked horizontal band if and only if $P_{hp}(y) > \eta_{hp} \forall y \in [y_{i} , y_{j}]$. Further, the vertical projection profile, $P_{vp}(x) = \sum_{y = y_{i}}^{y_{j}} \Omega(x,y) $ is calculated in every horizontal band bounded by $y_{i}$ and $y_{j}$; followed by thresholding with a threshold $\eta_{vp}$. A region between $x_{l}$ and $x_{k}$ in a particular horizontal band bounded by $y_{i}$ and $y_{j}$ is called text region if and only if $P_{vp}(x) > \eta_{vp}$, $\forall x \in [x_{l} , x_{k}]$.

Performance of the basic edge density based method is curtailed due to its high dependency on projection profile thresholds $\eta_{hp}$ and $\eta_{vp}$. Moreover, words are detected as text regions instead of text bands by the basic approach in \cite{Rosenfeld/Azriel:2003}, which is not favorable for TV broadcast news.

\begin{figure*}[htbp]
\centerline{
\fbox{\includegraphics[width = 0.5\textwidth]{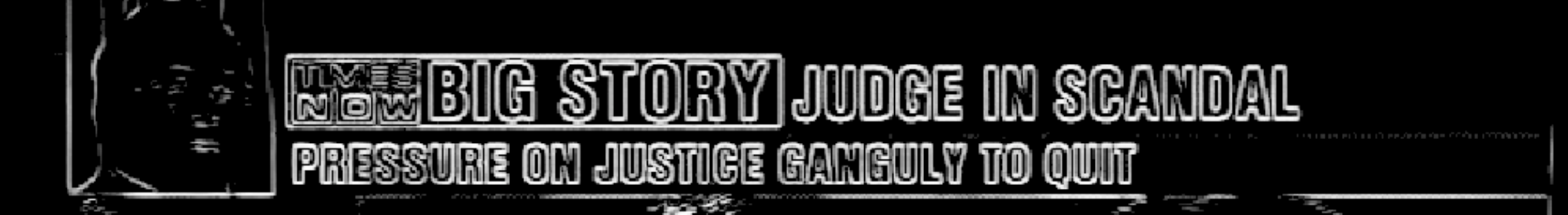}}
\fbox{\includegraphics[width = 0.16\textwidth]{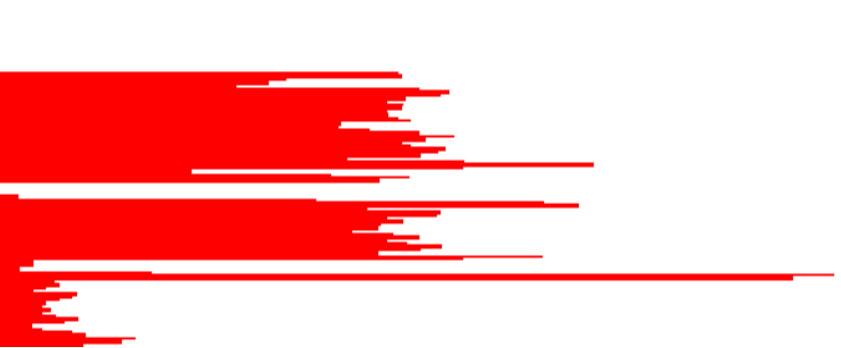}}
\fbox{\includegraphics[width = 0.16\textwidth]{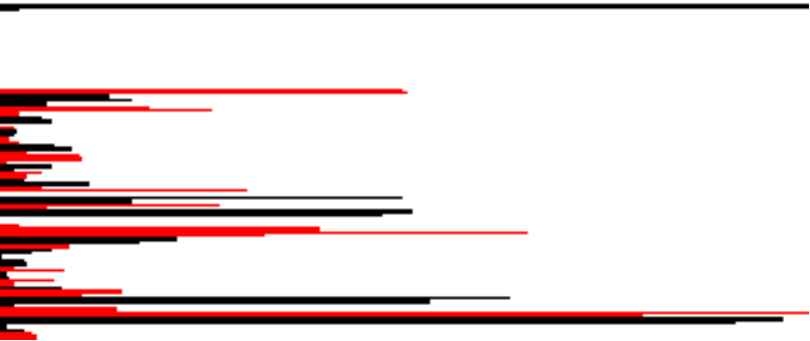}}
\fbox{\includegraphics[width = 0.16\textwidth]{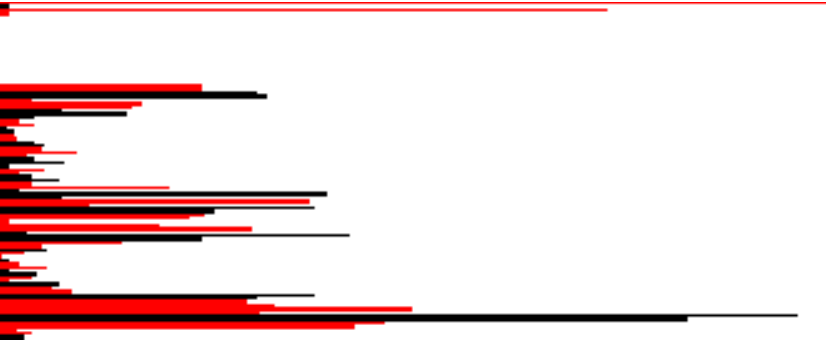}}
}
\centerline{\hspace{0.16\textwidth}(a) \hspace{0.32\textwidth} (b) \hspace{0.15\textwidth} (c)\hspace{0.15\textwidth} (d) }
\centerline{
\fbox{\includegraphics[width = 0.5\textwidth]{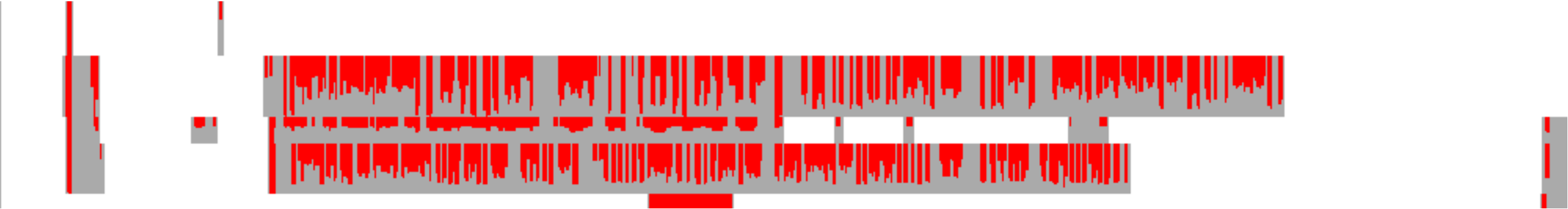}}
\fbox{\includegraphics[width = 0.52\textwidth]{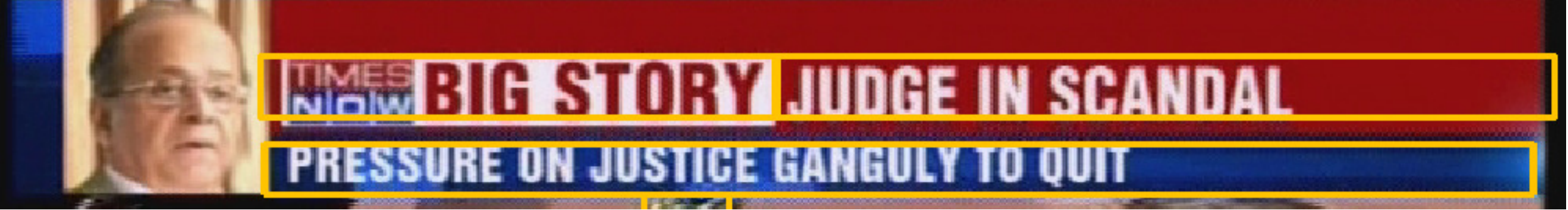}}
}
\centerline{ (e) \hspace{0.53\textwidth} (f)}
\caption{\footnotesize{Illustrating steps for text band localization. (a) is the modified gradient magnitude map. Text band boundaries are localized by using first (Fig.(c)) and second(Fig.(d)) differences of the horizontal projection profile (Fig. (b)). The red and black lines in (c) indicates the positive and negative differences respectively. The horizontal boundaries of text bands are located by finding the local extremum using second difference of projection profile. The vertical boundaries of text band are located by calculating the vertical projection profile of every located horizontal text band (Fig.(e)) (red lines represent magnitude of vertical profile). Localized text bands are shown in Fig.(f)}}
\vspace{-2em}
\label{fig:bndDet}
\end{figure*}

We propose to detect all possible horizontal lines first (arising due to text band boundaries) instead of horizontal bands. Once all these horizontal lines are located, we assume the region between each pair of consecutive lines as potential horizontal band and locate text regions by examining its vertical projection profile. Abrupt changes in the horizontal projection profile signifies the presence of a horizontal line or boundary. The first difference of the horizontal profile $P_{hp}^{'}(y)$, captures the abrupt changes in horizontal profile and has very high differences at boundary locations, while zero or very small differences elsewhere. Further, first difference of the profile have local extremum at boundary locations. Second difference $P_{hp}^{''}(y)$ of the horizontal projection profile locates these local extrema in first difference of the profile $P_{hp}^{'}(y)$ and hence, the potential text band boundaries. Various steps in text band detection are illustrated in Figure~\ref{fig:bndDet}.

The differentiated profile $P_{hp}^{'}$ (Figure~\ref{fig:bndDet}(b)) have non-zero values only at the locations where, discontinuities or horizontal lines are present. For a single horizontal line in image we will get multiple non-zero values in $P_{hp}^{'}$. An $\epsilon$-CCA (connected component analysis) is performed on $P_{hp}^{'}$ to group the prominent non-zero values (by assigning same label to horizontal lines in a group) arising due to single horizontal line. The result of CCA is stored in a label array $H_{l}$ where $L_{y}$ are the number of distinct labels assigned by CCA. This number of distinct labels is equal to the number of potential horizontal lines. The grouped non-zero values are shown marked in image~\ref{fig:bndDet}(b). The second difference of the projection profile $P_{hp}^{''}$ is thresholded by a local mean. The local mean is computed using set of points have same label in label array $H_{l}$. Identifying local minima in each region will give us the location of the horizontal line $H_{e}(l_{y})$ in respective local region.
\begin{eqnarray}
\displaystyle
\mu(l) = \frac{\sum_{i = 1}^{H}\left( |P_{hp}^{''}(i)| \delta_{k}( H_{l}(i) - l )  \right)}{\sum_{i = 1}^{H}\left( \delta_{k}( H_{l}(i) - l )  \right)}; l = 1,2..L_{y} \\
P_{hp}^{''}(y) = \begin{cases}
P_{hp}^{''}(y) & \text{$|P_{hp}^{''}(y)| > \mu\left(H_{l}(y)\right)$} \\
0 & \text{Otherwise}
\end{cases} ; y = 1,2,..H\\
H_{e}(l_{y}) = \left\{ y | \forall i \left[ P_{hp}^{''}(y) <  P_{hp}^{''}(i) \right] \wedge \left[ H_{l}(y) = l_{y} \right] \right\} 
\end{eqnarray}

Localized horizontal lines are shown in Figure~\ref{fig:bndDet}(c). Region between every consecutive pair of lines out of $L_{y}$ horizontal lines is considered to be a potential horizontal band. Vertical projection profile $P_{vp}^{l_{y}}$ is calculated in every potential horizontal band. Next, we use the $\epsilon-CCA$ (as done in case of horizontal profile) on first difference $P_{vp}^{'l_{y}}(x)$ of the vertical profile to locate the vertical band boundaries in each of the horizontal band. The Label array $V_{l}^{l_{y}}$ stores the result of CCA. The bounding boxes containing the text band are thus obtained from $P_{vp}^{'l_{y}}$  and $H_{e}(l_{y})$. 

The proposed approaches for image pre-processing (sub-section~\ref{subsec:preProc}) and text band detection have shown sufficient reduction of false positives in text band detection for individual frames (Table-\ref{tab:result}). We next introduce the framework for tracking (Section~\ref{sec:mtt}) the text bands detected in individual frames.

\section{Multiple Text Band Tracking}
\label{sec:mtt}

Tracking aims at associating the detection results across frames to establish a time history of the image plane trajectory of objects. This reduces the cost of detection and recognition in every video frame. Moreover, false detections arising out of local artifacts in some frames which do not persist for more than a few frames can be filtered out from trajectory analysis.
The multiple text region tracking is performed over two sets -- first, the set $\mathbf{tR}(\tau-1) = \{ tR_{i}(\tau - 1); i = 1,\ldots m_{\tau - 1} \}$ consisting of the text bounding rectangles $tR_{i}(\tau - 1)$ from the previous instant $\tau - 1$; and second, the set $\mathbf{dR}(\tau) = \{ dR_{i}(\tau); j = 1,\ldots n_{\tau} \}$ containing the bounding rectangles $dR_{j}(\tau)$ of the text regions detected in the present frame $I_{\tau}$.

The association between the text regions from two consecutive frames is measured by their overlap. For example, if $tR_{i}(\tau - 1)$ and $dR_{j}(\tau)$ has significant overlap, then we can conclude that both of them indicate the same text region and hence, the former can be updated by the later. However, errors in text detection do persist, either due to algorithm failure or on account of image quality. In many cases, either text bands appear fragmented after detection or the detection itself fails. This calls for an in-depth and formal analysis for identifying all such problem situations that the process of tracking may encounter. The associations between the elements of $\mathbf{tR}(\tau-1)$ and $\mathbf{dR}(\tau)$ are resolved in two steps. First, we construct the set $\mathbf{tS^{(d)}_{i}}(\tau) = \{ dR_{k}(\tau) : [ tR_{i}(\tau - 1)  \cup dR_{k}(\tau) \neq \Phi ] \}$ of detected regions overlapped with each of the previously tracked region  $\mathbf{tR}(\tau-1)$ and similarly, the set $\mathbf{dS^{(t)}_{j}}(\tau)  = \{ tR_{l}(\tau) : [ tR_{l}(\tau - 1)  \cup dR_{j}(\tau) \neq \Phi ] \}$ of previously tracked regions overlapped with each of the presently detected region $\mathbf{dR}(\tau)$. Next, we categorize the nature of these overlaps using the $RCC-5$ spatial relations.

The $i^{th}$ text region $tR_{i}(t)$ at the $t^{th}$ instant is represented by its bounding rectangle $bR_{i}(t)$ and its color histogram $cH_{i}(t)$. Thus, in ideal situations, a corresponding (detected) text region in the $t^{th}$ frame can be identified by checking the maximum overlap of $bR_{i}(t-1)$ and color match with $cH_{i}(t-1)$. However, problem cases like --  failure of text detection in the $t^{th}$ frame thereby losing correspondence; appearance of new text bands which have no association with previously tracked text bands; disappearance of existing text regions, call for the inclusion of reasoning through association analysis. The process of reasoning involves the estimation of qualitative spatial relations between text regions in $\mathcal{S}_{A}(t-1)$ and $\mathcal{S}_{D}(t)$. In this context, we explore the $RCC-5$ spatial relations and are described next.

\subsection{The RCC-5 Relations}
\label{subsec:rcc5}

Qualitative spatial relations are a common way to represent knowledge about configurations of interacting objects thereby reducing the burden of quantitative expressions of such relative arrangements. Besides, composition of existing relations allow the possibility of deducing newer relations among objects. There exists a vast literature on the spatial relations of overlapping objects, among which the $RCC-5$ region connection relations are most widely used. These relations consist of $DC$, $EQ$, $PO$, $PP$ and $PPi$. In the context of the estimation of these relations, we define the fractional overlap measure $\gamma_{fo}$ between two regions $A$ and $B$ as $\gamma_{fo}( A , B ) = {|A \cap B|}/{|A|}$. The predicates for detecting the object-blob $RCC-5$ relations by using the fractional overlap measure and with respect to a tolerance $\eta_{fo}$ are shown in table~\ref{tab:trk_detRCC5}. We next describe the procedure for identifying the different cases in the context of multiple text region tracking.
\begin{table}[!t]
\vspace{-1em}
\caption{\small{Detecting the $RCC-5$ relations using the thresholded fractional overlap measures.}}
\scriptsize
\vspace{-2em}
\begin{center}

\begin{tabular}{c|ccc}
\hline
\multirow{2}{*}{$\gamma_{fo}( A , B ) \downarrow \Big / \gamma_{fo}( B , A ) \rightarrow$}& \multirow{2}{*}{$\leq \eta_{fo}$}& \multirow{2}{*}{$\in ( \eta_{fo} , 1 - \eta_{fo} )$} & \multirow{2}{*}{$\geq ( 1 - \eta_{fo} )$} \\
\\\hline
$\leq \eta_{fo}$& $DC( \eta_{fo} )$& $PO( \eta_{fo} )$& $PPi( \eta_{fo} )$ \\ 
$\in ( \eta_{fo} , 1 - \eta_{fo} )$	& $PO( \eta_{fo} )$& $PO( \eta_{fo} )$ & $PPi( \eta_{fo} )$\\
$\geq 1- \eta_{fo}$& $PP( \eta_{fo} )$& $PP( \eta_{fo} )$ & $EQ( \eta_{fo} )$ \\ \hline
\end{tabular}
\end{center}
\label{tab:trk_detRCC5}
\vspace{-3em}
\end{table}

\textbf{Unique Correspondence} -- A text region $tR_{i}(t-1)$ is considered to have an unique correspondence with a detected text region $dR_{j}(t)$, if $|\mathbf{tS^{(d)}}_{i}(\tau)| = 1 $ and $|\mathbf{dS^{(t)}}_{j}(\tau)| = 1 $. In this case, $tR_{i}(t-1) \in \mathcal{S}_{A}(t-1)$ is updated with the bounding rectangle and color histogram of $dR_{j}(t)$ to form $tR_{i}(t) \in \mathcal{S}_{A}(t)$. The update rule for updating the bounding box and color histogram is determined by RCC-5 Relation between two regions. Four possible relations are shown in the first row of table \ref{tab:trkCases}. The fluctuations in localizing the band boundary in detection stage are handled by these conditions.

\textbf{Multiple Correspondences} can have three different types -- First, multiple tracked text regions $tR_{i}(t-1)$ can uniquely overlap with a single detected text band $dR_{j}(t)$ i.e.$(|\mathbf{tS^{(d)}}_{i}(\tau)| = 1) $ and $(|\mathbf{dS^{(t)}}_{j}(\tau)| > 1)$, $ \forall i tR_{i}(t - 1) \cup dR_{j}(t) \neq \Phi $. Second, multiple detected text regions $dR_{j}(t)$ can uniquely overlap with a single tracked band $tR_{i}(t-1)$, i.e. $(|\mathbf{tS^{(d)}}_{i}(\tau)| > 1)$ and $(|\mathbf{dS^{(t)}}_{j}(\tau)| = 1) $ ,$\forall j tR_{i}(t - 1) \cup dR_{j}(t) \neq \Phi$. Third, multiple tracked text regions $tR_{i}(t-1)$ can overlap with multiple detected text regions, i.e. $(|\mathbf{tS^{(d)}}_{i}(\tau)| > 1)$  and $(|\mathbf{dS^{(t)}}_{j}(\tau)| > 1)$, $ \forall i,j tR_{i}(t - 1) \cup dR_{j}(t) \neq \Phi$.

The possible cases are listed in table~\ref{tab:trkCases}. First case arises due to multiple track initializations on same text band. The possibility of merging is checked and the bounding boxes are updated accordingly. Second case occurs when a tracker was initialized on group of text regions due to detection failures. The tracked band is checked for splitting according to newly detected regions. In the third case, there is a possibility for both splitting as well as merging and both are checked.

\textbf{Disappearance} -- A text band $tR_{i}(t-1)$ is considered to disappear in the next frame if it does not have any correspondence with any detected text band in $\mathcal{S}_{D}(t)$, i.e. $\forall j bR_{i}(t-1)~DC( \eta_{fo} )~dR_{j}(t)$ or $(|\mathbf{tS^{(d)}}_{i}(\tau)| =  1)$. This may be due to actual termination of the display of that text or detection failure due to local artifacts in present frame.

In this case, we first focus on the bounding rectangle box $bR_{i}(t-1)$ of $tR_{i}(t-1)$ in the current frame. If the color histogram obtained from $bR_{i}(t-1)$ in the current frame matches with that of $cH_{i}(t-1)$ then, we call it a temporary detection failure and restore the track of $tR_{i}(t-1)$. Otherwise, we consider this as the termination of display of the tracked text and do not include it further in $\mathcal{S}_{A}(t)$.

\textbf{New Entry} -- The detected text region $dR_{j}(t)$ is considered to be a new entry if it does not have any correspondence with the text bands in $\mathcal{S}_{A}(t-1)$, i.e. $\forall i bR_{i}(t-1)~DC~dR_{j}(t)$ or $ (|\mathbf{dS^{(t)}}_{j}(\tau)| = 0)$. In this case, we form a new text region with the bounding rectangle and color histogram of $dR_{j}(t)$ and include it in $\mathcal{S}_{A}(t)$.

The system initializes with $\mathcal{S}_{A}(0) = \Phi$. The first set of detections are inducted in the set $\mathcal{S}_{A}$ and the tracking process continues with addition, update and removal of text bands by association analysis of tracked and detected text bounding rectangles. Each tracked text region have the background and foreground information. This color information is further used to binarize the text bands. These binarized text bands are then accumulated over the entire track before passing to the OCR engine so as to save multiple passes of OCR. Next, we describe the modifications made on Tesseract OCR to improve the text recognition.

\begin{table}[!t]
\caption{\small{Possible cases of association between the tracked and detected rectangles of text regions}}
\scriptsize
\vspace{-2em}
\begin{tabular}{cc|cccc}\\ \hline
\multirow{5}{*}{\rotatebox[origin=c]{90}{$|\mathbf{tS^{(d)}}_{i}(\tau)|$}}& \multirow{5}{*}{\rotatebox[origin=c]{90}{$|\mathbf{dS^{(t)}}_{j}(\tau)|$}}  &\multirow{5}{0.15\linewidth}{$tR_{i}(\tau-1)$ $\{EQ\}$ $dR_{j}(\tau)$}  &\multirow{5}{0.15\linewidth}{$tR_{i}(\tau-1)$ $\{PP\}$ $dR_{j}(\tau) $}  & \multirow{5}{0.15\linewidth}{$tR_{i}(\tau-1)$ $\{PPi\}$ $dR_{j}(\tau) $} & \multirow{5}{0.15\linewidth}{$tR_{i}(\tau-1)$ $\{PO\}$ $dR_{j}(\tau)$}  \\
 &&\\
 &&\\
&&\\
&&\\\hline
$1$ & $1$ &\includegraphics[width=0.125\columnwidth]{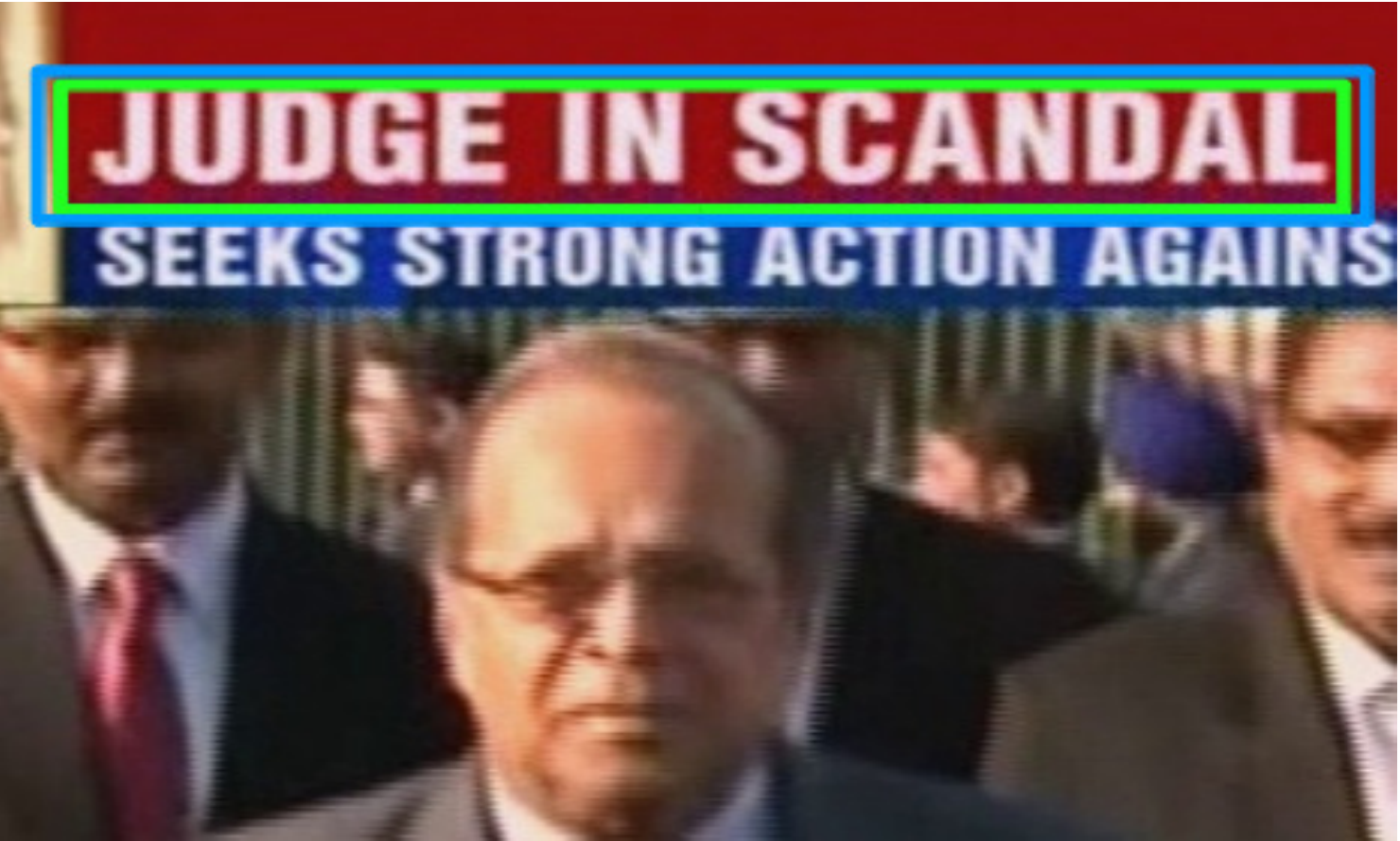}&\includegraphics[width=0.125\columnwidth]{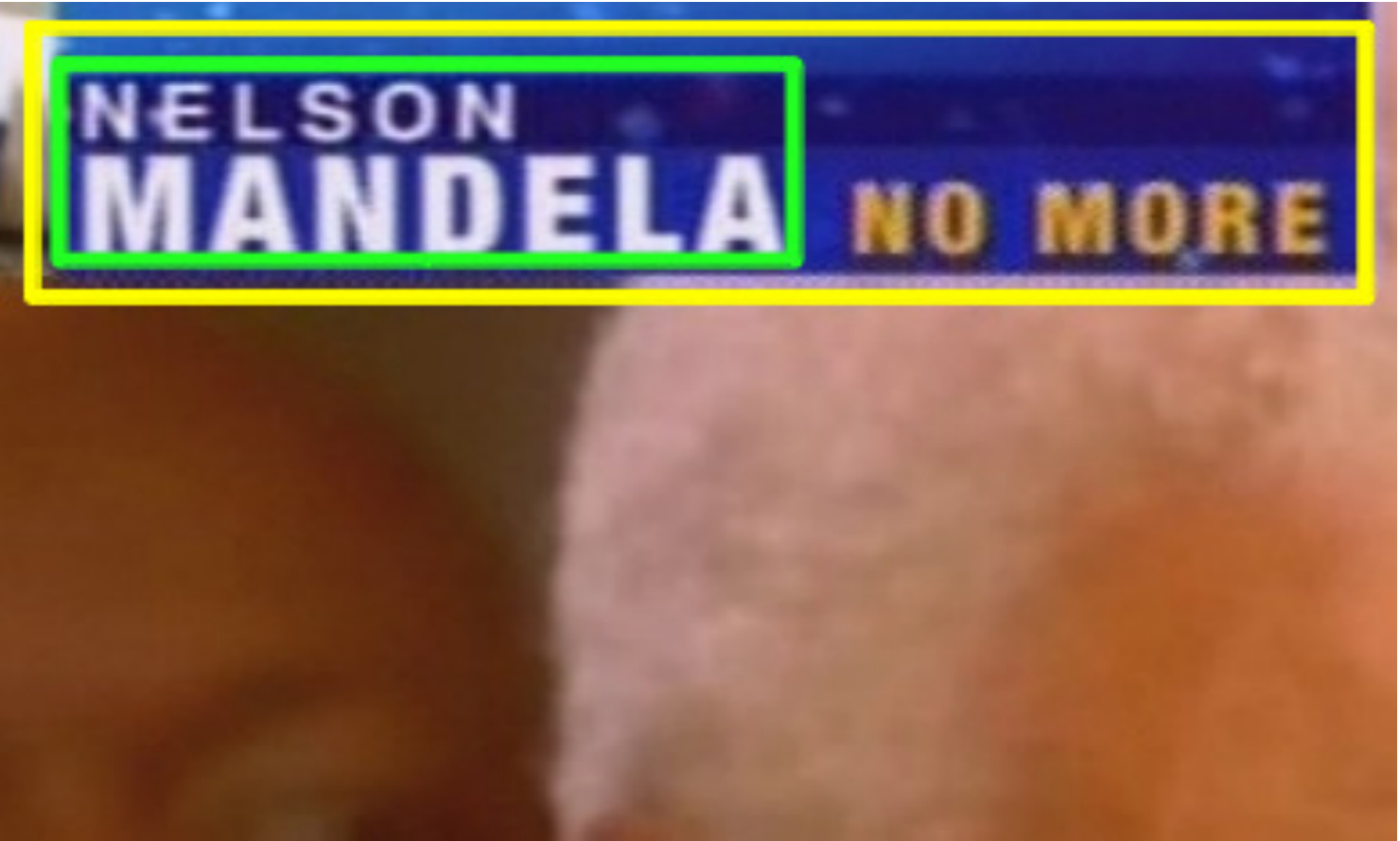}&\includegraphics[width=0.125\columnwidth]{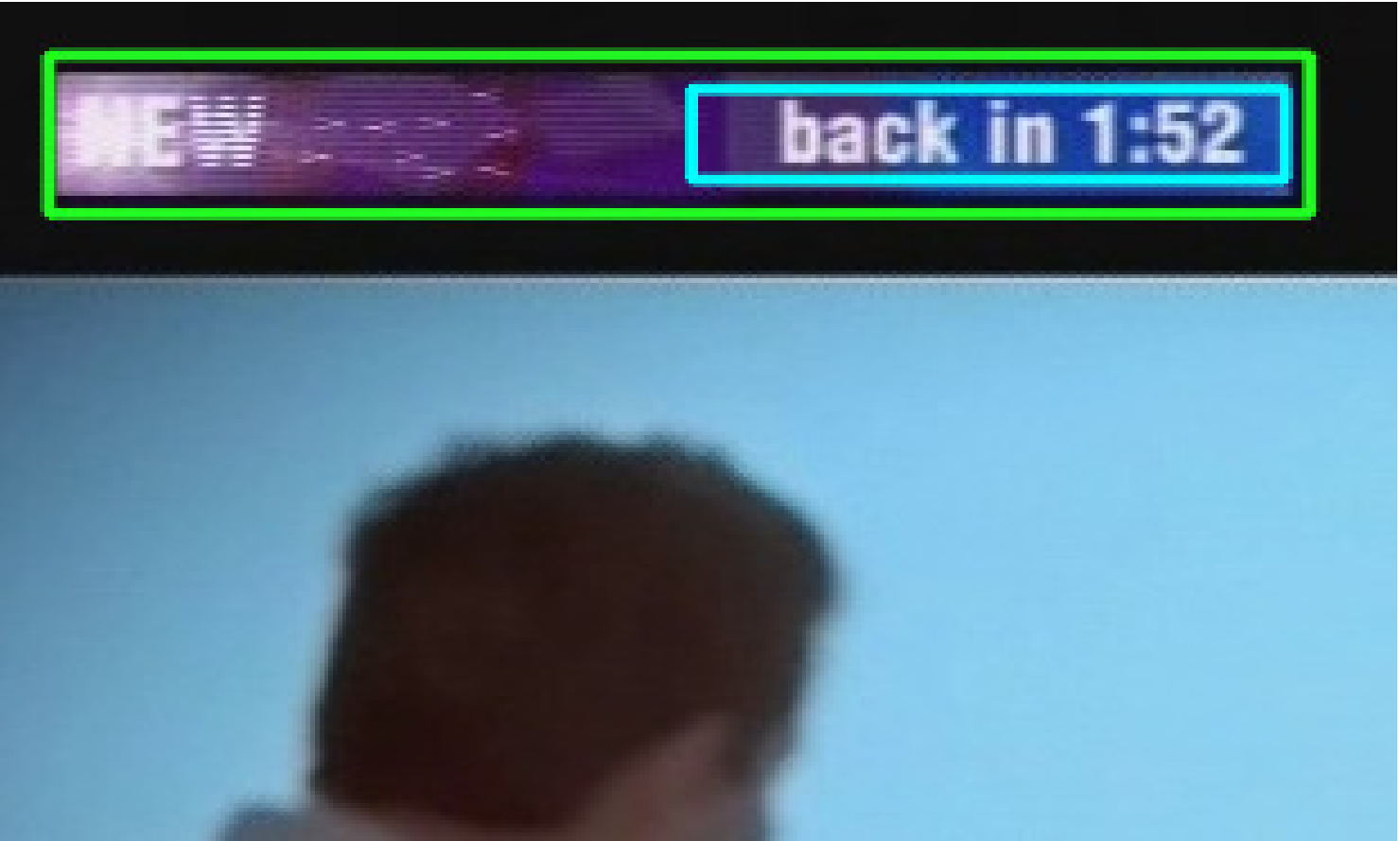}&\includegraphics[width=0.125\columnwidth]{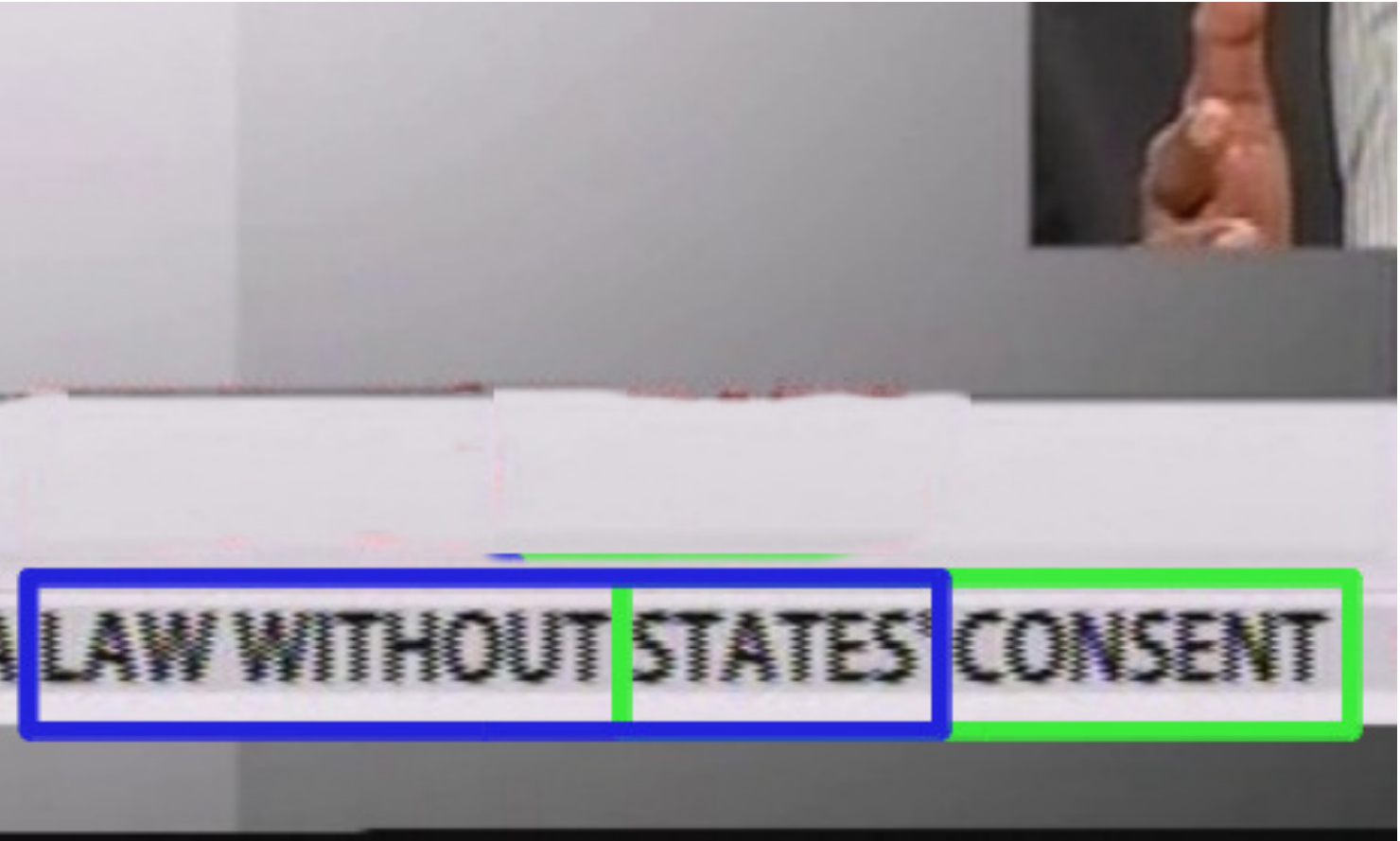} \\ \hline
$1$ & $1+$ & & & \includegraphics[width=0.125\columnwidth]{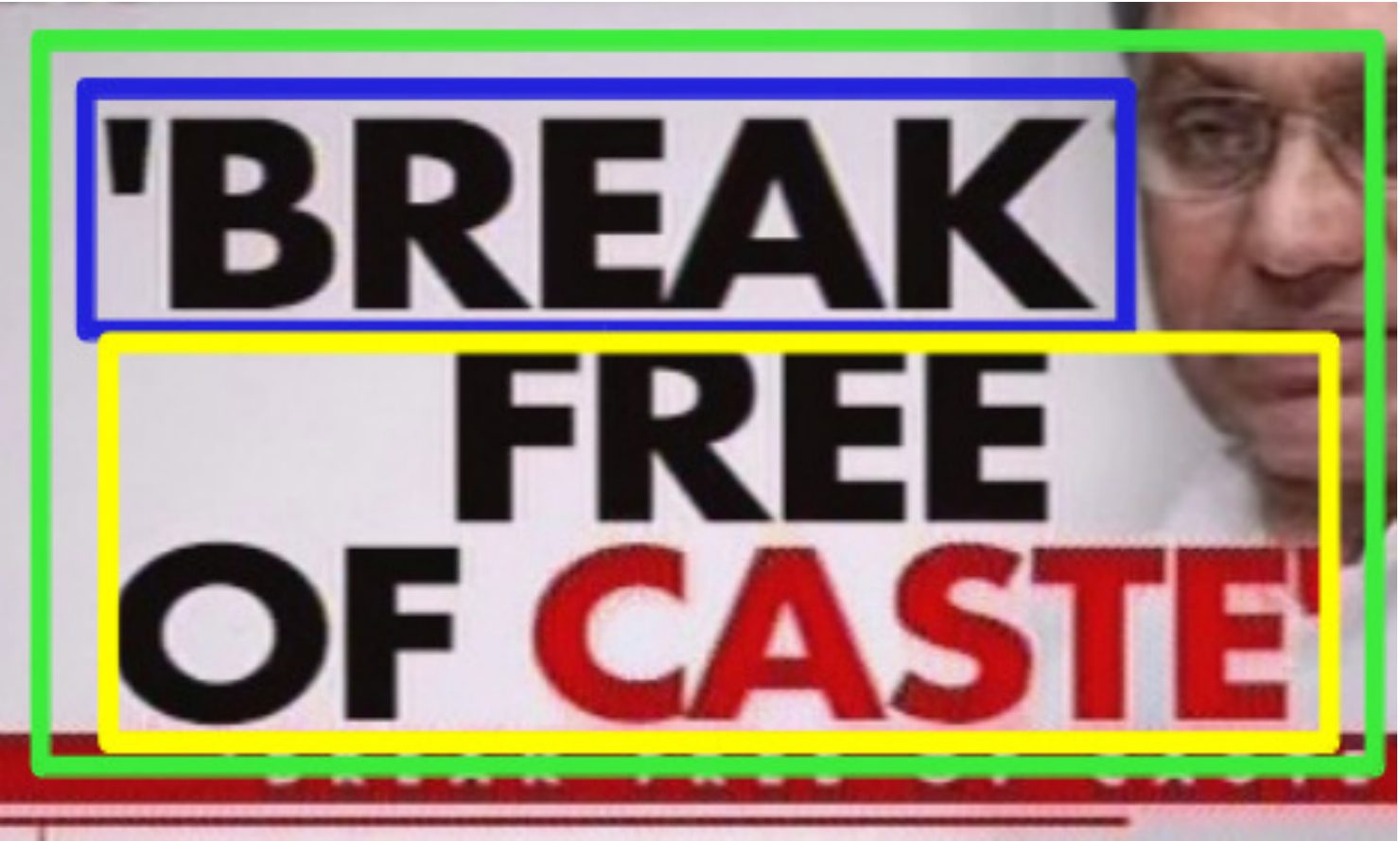}  & \includegraphics[width=0.125\columnwidth]{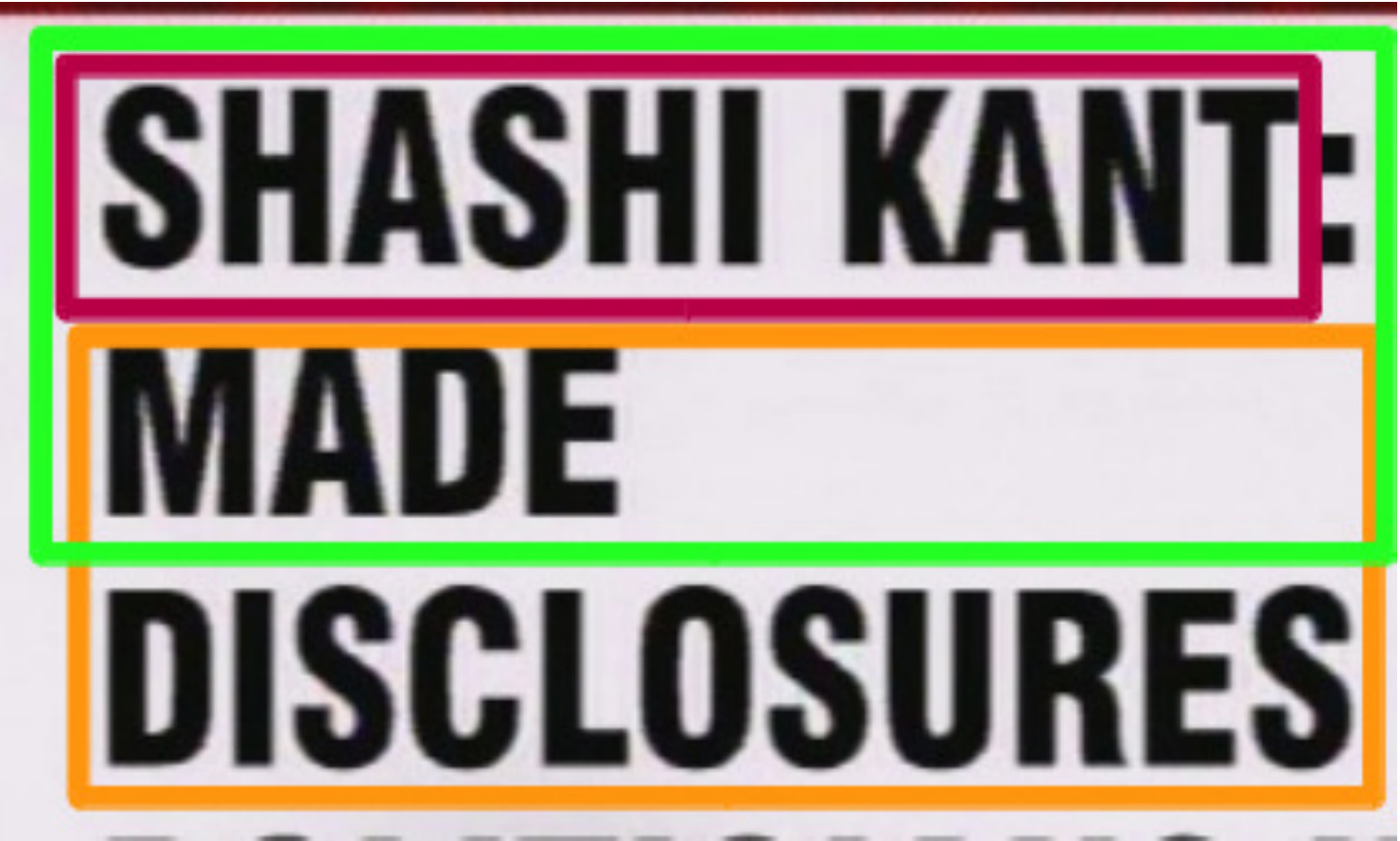} \\ \hline
$1+$ & $1$ & & \includegraphics[width=0.125\columnwidth]{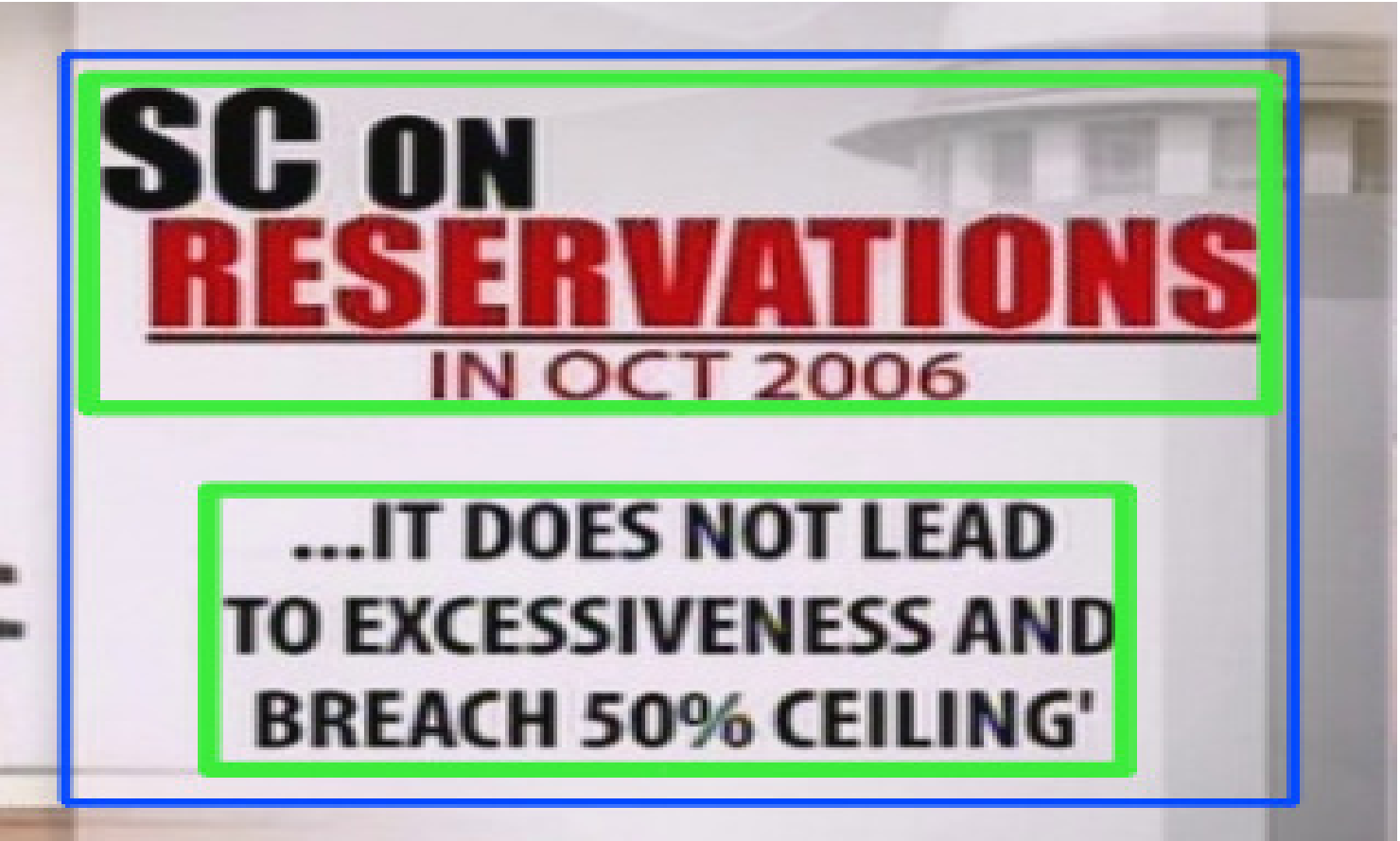} &   & \includegraphics[width=0.125\columnwidth]{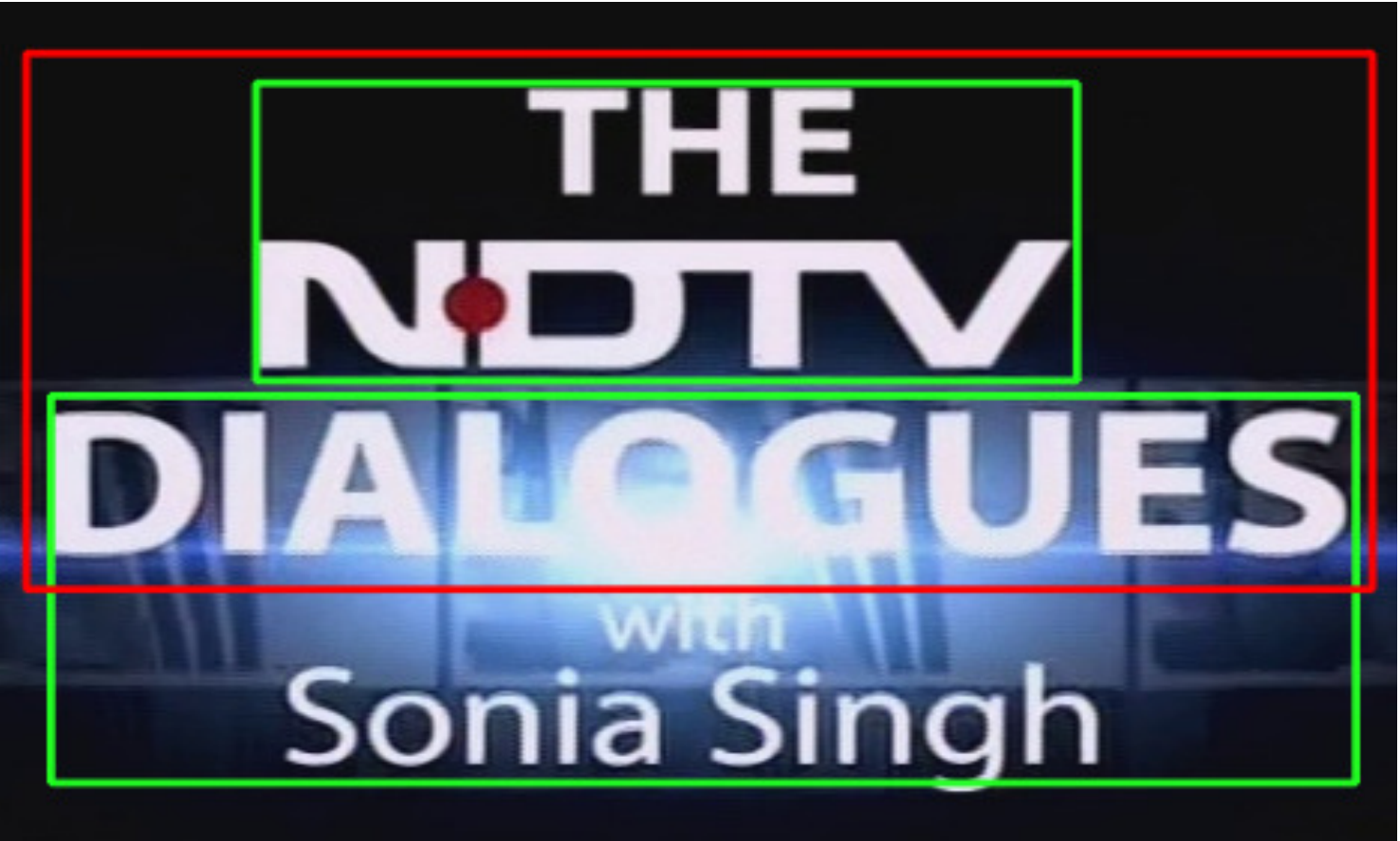} \\ \hline
$1+$ & $1+$ & &  & & \includegraphics[width=0.125\columnwidth]{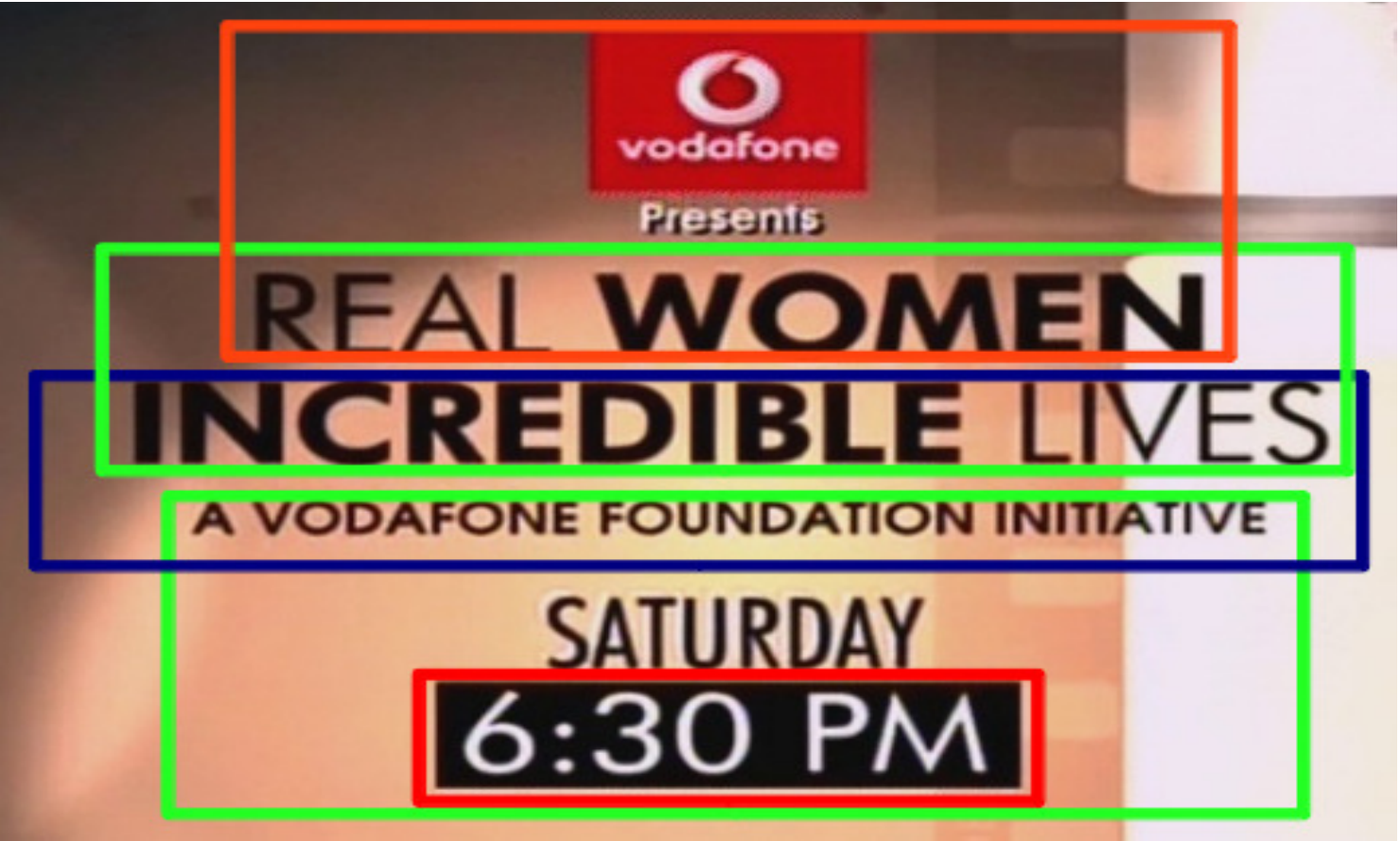} \\ \hline
\end{tabular}
\label{tab:trkCases}
\vspace{-2em}
\end{table}

\section{Adapting Tesseract For Overlaid Text Recognition}
\label{sec:ocr}

Tesseract \cite{tesseract} OCR engine was primarily developed for optical character recognition on scanned documents at HP labs during $1984$ and $1994$. In late $2005$, HP released Tesseract for open source. Tesseract provides basic OCR engine with tools for training. We have used Tesseract OCR on aggregated binarized images of consistently tracked bands to recognize the text content. OCR engine uses adaptive character segmentation based on linguistic model and character classifier. Font dependent polygonal approximation of the character outline is used as a feature for the character classifier. Therefore, linguistic model and the fonts used for training the character classifier plays crucial role in overall performance of Tesseract. Tesseract OCR engine is by default trained for scanned documents and uses standard English language model. Whereas, overlaid text often contains proper nouns and hash tags presented in variety of fonts. Hence, Tesseract with default models performs poorly on overlaid text recognition. To adapt the OCR engine for the task of overlaid text recognition it is necessary to train the character classifier with overlaid text samples and update the language model to include proper nouns and hash tags.

First, for training character classifiers sufficient ground truth data is required. Marking character level ground truth data for text recognition is a very tedious job. In literature, it is argued that polygonal approximation feature used in Tesseract is only font dependent. Hence, instead of marking the character level ground truth data, we have identified the commonly used overlaid text fonts (we have used $34$ fonts) across channels and synthetically generated the data for training the Tesseract. 
Second, in order to enrich the linguistic model used during recognition, we have used the news articles available on various news websites. News articles are rich sources of proper nouns, hash tags (present in meta data) and abbreviations. We have used a corpus of approximately $13,00,000$ web articles along with meta data provided on websites (like hash tags, tweets) from three different sources viz. NDTV, Times of India and First Post published during January, $2011$ to April, $2015$. Tesseract requires a dictionary of words, list of bi-grams, list of frequently occurring words, list of numbers and list of punctuations, as directed acyclic word graphs to build the linguistic model. From the corpus of web articles, all required  directed acyclic word graphs are generated. In the next section, we present the results of experimentation. 
%
\section{Results}
\label{sec:res}
We have evaluated the performance of each sub-system separately. We have tested our preprocessing (CE) scheme with stroke width transform~\cite{SWT:CVPR2010} (SWT) and projection profile based method (\cite{Rosenfeld/Azriel:2003})(PP), on $3$ different image datasets of two different types viz. natural scenes and born-digital images. Though the preprocessing method is primarily developed for overlaid text, the assumption of high contrast of text regions holds at large for natural scene text as well. Hence, we have experimented on natural scene images as well as born digital images. The natural scene images are from ICDAR $2003$ ($509$ images, test and training set) and ICDAR $2013$ ($452$ images, test and training set) and Born-digital images are from ICDAR $2011$ ($420$ images). The performance of text detection/localization with and without preprocessing is compared using precision, recall and f-measure, calculated by \emph{Ephstein's Criterion} \cite{SWT:CVPR2010}. SWT works on binary image and hence, histogram equalized gradient magnitude image is thresholded and non-maximal suppression is used to obtain the binary image. Both the text detection algorithms have shown improvement in the performance with the proposed preprocessing method (Table-\ref{tab:result}). 

\begin{table}[htbp]
\vspace{-2em}
\caption{\small{Performance analysis of pre-processing and text band detection}}
\centering
\scriptsize
\vspace{-1em}
\begin{tabular}{|c|c|l|l|l|l|}
\hline
\multicolumn{ 1}{|c|}{Datasets} & \multicolumn{ 1}{c|}{Method} & \multicolumn{ 3}{c|}{Epshtein's Criterion} & \multicolumn{ 1}{c|}{Time} \\ \cline{ 3- 5}
\multicolumn{ 1}{|c|}{} & \multicolumn{ 1}{c|}{} & \multicolumn{1}{c|}{P} & \multicolumn{1}{c|}{R} & \multicolumn{1}{c|}{F} & \multicolumn{ 1}{l|}{(Sec)} \\ \hline
& SWT & 0.5122 & 0.6086 & 0.5563 & 0.97 \\
& CE+SWT & \textbf{0.5142} & \textbf{0.7106} & \textbf{0.5967} & 1.01 \\
& PP & 0.2958 & 0.2765 & 0.2856 & 0.063 \\
\multirow{-4}{*}{\rotatebox[origin=c]{45}{ICDAR 2013}} & CE+PP & 0.3314 & 0.4026 & 0.3636 & 0.081 \\ \hline
& SWT & 0.3829 & 0.4296 & 0.4049 & 0.217 \\
& CE+SWT & 0.3995 & 0.4394 & 0.4185 & 0.225 \\
& PP & 0.2935 & 0.4954 & 0.3689 & 0.043 \\
\multirow{-4}{*}{\rotatebox[origin=c]{45}{ICDAR 2011}}& CE+PP & \textbf{0.382} & \textbf{0.4864} & \textbf{0.4282} & 0.054 \\ \hline
& SWT & 0.579 & 0.6299 & 0.6034 & 0.829 \\
& CE+SWT & \textbf{0.611} & \textbf{0.7053} & \textbf{0.6548} & 0.87 \\
& PP & 0.5012 & 0.4853 & 0.4931 & 0.059 \\
\multirow{-4}{*}{\rotatebox[origin=c]{45}{ICDAR 2003}}& CE+PP & 0.5868 & 0.6605 & 0.6215 & 0.063 \\ \hline
Our Dataset&SWT & 0.6502 & 0.6894 & 0.6692 & 0.85 \\
Text Band&CE+SWT & 0.7292 & 0.7826 & 0.755 & 0.907 \\
Detection&PP-TB & 0.5327 & 0.6108 & 0.5691 & 0.061 \\
&CE+PP-TB & \textbf{0.76} & \textbf{0.8544} & \textbf{0.8045} & 0.084 \\ \hline
\end{tabular}
\label{tab:result}
\vspace{-2em}
\end{table}

The performance of text band localization is evaluated on our own dataset having $150$ ($720 \times 576$) images containing challenging detection cases in news videos. For our dataset, the ground truth is marked for band detection instead of word detection. On our dataset of news video images, projection profile based text band detection (PP-TB) outperforms the other methods (Table \ref{tab:result}).

The drop in recall of SWT is due to the poor quality of the video frames. More so, the proposed method is a lot faster than SWT and other reported methods. While evaluating the performance of SWT on our dataset to compensate for difference in ground truth, we relax the matching criterion to favor SWT. Our implementation of SWT took an average time of $0.85$ seconds for an image size of $720 \times 576$ (reported time is $0.90$ sec for $ 640 \times 480$ image) while our approach took an average time of $84$ msecs only.

The performance of tracking is evaluated by using the purity of the track as well as track switches. A track is called pure when it has single properly tracked text band. While track switch occurs when multiple tracks are generated on a single ground truth track. We have evaluated the tracking performance on $3$ videos of $1$ hour each from $3$ different Indian news channels. In all, we obtained $1,05,630$ tracks out of which $73,053$ tracks were found to be pure with $10,457$ track switches. However remaining $22,120$ tracks were either initialized wrongly or tracked incorrectly. 

We have evaluated the performance of OCR engine on three hours of video data from three different channels. The word level and character level error rates are presented in table~\ref{tab:ocr}. It is evident from the table that adapting Tesseract using web articles has improved performance of the OCR significantly.

\section{Conclusion}
\label{sec:conc}

We have proposed a methodology for overlay text extraction in TV broadcast news videos. In the context of text detection and localization, we have significantly improved over existing edge density based methods. We observed that this basic approach had high false positives on account of strong edges present in non-text regions. We have proposed a threshold free preprocessing scheme for suppression of non-text edges while boosting text edges. The effects of stronger edges coming from non-text regions are nullified further by using the first and second order derivatives of the edge density projection profiles while localizing the text bands. The detected text regions are tracked across the frames to extract the static and consistent text bands using a formal reasoning framework. The use of RCC-5 based reasoning framework allowed us to identify different problem cases in detection and tracking. Finally, text is extracted from consistently tracked text bands using Tesseract OCR engine trained using web news articles. All our proposals have shown significant improvement in the performance of each subsystem.
\begin{table}[htbp]
\vspace{-2em}
\caption{Table shows the character and word level accuracies of the OCR engine. The word level performance is reported after applying dictionary corrections. Smaller errors are better}
\small
\vspace{-2em}
\begin{center}

	\begin{tabular}{cccccc}
	\\\hline
&  \multicolumn{2}{c}{Character Level} & &  \multicolumn{2}{c}{Word Level} \\ \cline{2-3} \cline{5-6}
& Errors & \%Error & &Errors & \%Error \\ \hline
Tesseract-Default &67167 &20.97 & &60805 &56.94 \\
Tesseract-Modified &\bf 16016 &\bf4.99& &\bf7527 & \bf7.04 \\ \hline
\end{tabular}
\end{center}
\label{tab:ocr}
\vspace{-2em}
\end{table}
This work was a first step towards the larger goal of broadcast (news) video analytics. The textual data acts as an important source of information for obtaining meaningful tags for displayed events, persons (faces), places and time. Thus, the extracted text data can be used further as features for applications like video event classification, news summarization, news story segmentation and association of names to places and persons.

\section{Acknowledgement}
This work is part of the ongoing project on ``Multi-modal Broadcast Analytics: Structured Evidence Visualization for Events of Security Concern'' funded by the Department of Electronics \& Information Technology (DeitY), Govt. of India.


{
\tiny
\bibliographystyle{IEEEbib}
\bibliography{1570185507}
}

\end{document}